\documentclass[runningheads]{llncs}
\usepackage[T1]{fontenc}
\usepackage{amsmath}
\usepackage{array}
\usepackage{xspace}
\usepackage{graphicx}
\usepackage{hyperref}
\usepackage{multirow}
\usepackage{subcaption}
\usepackage{booktabs, boldline}
\usepackage{arydshln}

\usepackage{color}

\newcommand{\modeladvmultvae}{\texttt{\textsc{AdvMultVAE}}\xspace}
\newcommand{\modelmadvmultvae}{\texttt{\textsc{AdvXMultVAE}}\xspace}
\newcommand{\modelmultvae}{\texttt{\textsc{MultVAE}}\xspace}
\newcommand{\modeladvmultvaegen}{\texttt{\textsc{AdvMultVAE-G}}\xspace}
\newcommand{\modeladvmultvaeage}{\texttt{\textsc{AdvMultVAE-A}}\xspace}
\newcommand{\modelmadvmultvaegen}{\texttt{\textsc{AdvXMultVAE-G}}\xspace}
\newcommand{\modelmadvmultvaeage}{\texttt{\textsc{AdvXMultVAE-A}}\xspace}

\newcommand{\dataMlm}{\texttt{\textsc{Ml-1m}}\xspace}
\newcommand{\dataLFM}{\texttt{\textsc{LFM-2b-100k}}\xspace}
\newcommand{\dataorigLFM}{\texttt{\textsc{LFM-2b}}\xspace}

\newcommand{\lambgender}{$\lambda_{Gender}$}
\newcommand{\lambage}{$\lambda_{Age}$}
\newcommand{\baccgender}{$BAcc_{G}$\xspace}
\newcommand{\maeage}{$MAE_{A}$\xspace}

\newcommand{\eg}{e.\,g., }

\newcommand{\ndcg}{\text{NDCG}\xspace}

\newcommand{\Recall}{\text{Recall}\xspace}

\def\Tabref#1{Table~\ref{#1}}

\begin{document}
\title{Simultaneous Unlearning of Multiple Protected User Attributes From Variational Autoencoder Recommenders Using Adversarial Training}
\titlerunning{Simultaneous Unlearning of Multiple Attributes}
%
\author{
Gustavo Escobedo\inst{1}\orcidID{0000-0002-4360-6921} \and
Christian Ganh{\"o}r\inst{1}\orcidID{0000-0003-1850-2626} \and
Stefan Brandl\inst{1}\orcidID{0000-0001-5254-3005} \and
Mirjam Augstein\inst{3}\orcidID{0000-0002-7901-3765} \and
Markus Schedl\inst{1,2}\orcidID{0000-0003-1706-3406}
}
\authorrunning{G. Escobedo et al.}
%
\institute{
Johannes Kepler University Linz, Linz, Austria\\
\and
Linz Institute of Technology, Linz, Austria\\
\email{\{gustavo.escobedo,christian.ganhoer,stefan.brandl,markus.schedl\}@jku.at}\\
\and
University of Applied Sciences Upper Austria, Hagenberg, Austria\\
\email{mirjam.augstein@fh-hagenberg.at}
}
\maketitle              
\begin{abstract}
In widely used neural network-based collaborative filtering models, users' history logs are encoded into latent embeddings that represent the users' preferences. In this setting, the models are capable of mapping users' protected attributes (e.g., gender or ethnicity) from these user embeddings even without explicit access to them, resulting in models that may treat specific demographic user groups unfairly and raise privacy issues. While prior work has approached the removal of a single protected attribute of a user at a time, multiple attributes might come into play in real-world scenarios.    
In the work at hand, we present \modelmadvmultvae which aims to unlearn multiple protected attributes (exemplified by gender and age) simultaneously to improve fairness across demographic user groups. For this purpose, we 
couple a variational autoencoder (VAE) architecture with adversarial training (\modeladvmultvae) to support simultaneous removal of the users' protected attributes with continuous and/or categorical values. Our experiments on two datasets, \dataLFM and \dataMlm, from the music and movie domains, respectively, show that our approach can yield better results than its singular removal counterparts  (based on \modeladvmultvae) in effectively mitigating demographic biases 
whilst improving the anonymity of latent embeddings.

\end{abstract}
\section{Introduction}

Recommender system models commonly leverage collaborative information in order to predict the likelihood of an item to be consumed by a user. Most neural network-based recommender models, including \modelmultvae~\cite{LiangMultVAE2018},  encode the users' consumption histories into latent vector representations which are used to produce recommendations. Though these latent vectors are based only on user consumption data, they can implicitly encode undesirable biases and sensitive user attributes, which are a potential source of unfairness and privacy issues, respectively \cite{10.1145/3511047.3536400,MELCHIORRE2021102666}. As a result, users from specific demographic groups may be treated unfairly regarding the quality and/or diversity of recommendations. Furthermore, users' protected information (e.g.,  gender, ethnicity, age) is subject to potential exposure through adversarial attacks~\cite{DBLP:journals/csur/DeldjooNM21}, which directly harms the trustworthiness of recommender models~\cite{Wang2023TrustRec}.  

To tackle these issues, prior work \cite{10.1145/3477495.3531820} applied adversarial information removal techniques in order to remove the users' gender information from latent vectors while maintaining similar recommendation performance. However, in real-world scenarios, additional protected attributes such as age or country might be encoded in latent space at the same time. Therefore, in this work, we introduce \modelmadvmultvae that aims at jointly removing several protected attributes by extending the \modeladvmultvae architecture~\cite{10.1145/3477495.3531820} to support simultaneous removal of user-protected attributes (continuous and/or categorical). For our experiments, we leverage two publicly available datasets that include protected attributes: a subset of \dataorigLFM\footnote{\url{http://www.cp.jku.at/datasets/LFM-2b/}}, which we name \dataLFM, and \dataMlm\footnote{\url{https://grouplens.org/datasets/movielens/1m/}} from the music and movie domains,  respectively. 
We evaluate the proposed \modelmadvmultvae against \modeladvmultvae by studying
the removal of gender (categorical) and age  (continuous) attributes under different degrees of removal aggressiveness. Our results indicate that \modelmadvmultvae can outperform or are on-par with the baselines in terms of gender and age removal while maintaining competing performance in terms of \ndcg and \Recall. 
\section{Related Work}
Adversarial training techniques have been used for removing protected user attribute information from latent embeddings to learn fair and privacy-preserving representations in classification and retrieval tasks~\cite{Wang23SurveyFairness,hauzenberger-etal-2023-modular,kumar-etal-2023-parameter,beigi2020survey}.   
Also recommendation models are not exempt of privacy-related threats \cite{DBLP:journals/csur/DeldjooNM21}. Therefore, prior work introduced several mitigation approaches. For instance, Wang et al.~\cite{Wang2022InferenceAttack} presented a VAE-based disentangled encoder and an attacker network to avoid membership inference attacks by removing biases from training data. Furthermore, Liu et al.~\cite{adv-LiuWLXZ22} applied removal of gender to mitigate user data exposure and improve fairness in GNN-based recommenders. Also, Li et al.~\cite{Li2021counterFactMult} proposed an adversarial learning approach to generate filters for user embeddings to remove different combinations of categorical user protected attributes to improve fairness in matrix-factorization based recommenders.
In addition, Ganh\"or et al.~\cite{10.1145/3477495.3531820} introduced \modeladvmultvae in order to unlearn users' gender information from latent embeddings while maintaining recommendation accuracy. In this work, we also focus on learning privacy-preserving representations in VAE recommenders, but different from previous approaches where users' protected attributes are treated one at a time, we perform simultaneous attribute removal. 

\section{Method}\label{sec:method}

We briefly introduce the \modelmultvae and \modeladvmultvae models and their components followed by the description of our proposed \modelmadvmultvae architecture. 

The \modelmultvae~\cite{LiangMultVAE2018} is a recommendation model that consists of two sub-networks, namely, an encoder $f(\cdot)$ and a decoder $g(\cdot)$. The encoder transforms the users' implicit preferences vector $x$ into the latent distribution $\mathcal{N}(\mu,\sigma)$ characterized by the learnable parameters $\mu$ and $\sigma$ using a Gaussian prior $\mathcal{N}(0,I)$  from which the latent vector $z$ is sampled. 
The vector $z$ serves as input to the decoder sub-network $g(\cdot)$, generating $g(z)$ which is used to reconstruct the original users' implicit preferences vector $x$ through the minimization of the reconstruction loss $\mathcal{L}^{REC}(g(z),x)$ and the regularization term $\mathcal{L}^{KL}(\mathcal{N}(\mu,\sigma),\mathcal{N}(0,I))$, defined as the Kullback-Leibler divergence between the estimated latent distribution given by the parameters $\mu$ and $\sigma$ and its prior. The resultant loss $\mathcal{L}^{MULT}$ is shown in Equation~\ref{eq:lossmultvae} where the hyperparameter $\beta$ is introduced as a factor to adjust of regularization power of $\mathcal{L}^{KL}$ which steers the learning to its prior.

\begin{equation}
    \label{eq:lossmultvae}
    \mathcal{L}^{MULT} = \mathcal{L}^{REC}(g(z),x) - \beta\mathcal{L}^{KL}(\mathcal{N}(\mu,\sigma),\mathcal{N}(0,I))
\end{equation}

The \modeladvmultvae~\cite{10.1145/3477495.3531820} model is an extension of  \modelmultvae and is composed of two training phases: an adversarial training phase to remove sensitive information from the latent vectors and an adversarial attack phase aiming to recover remaining information. For the adversarial training, an adversarial network module $h(\cdot)$ is added as an extra prediction head plugged into the output of the encoder $f(\cdot)$. Its main task is to predict the users' protected attribute $y$ from the latent vector $z$ while maintaining recommendation performance. Consequently, the learning process can be defined as a min-max problem where  $\mathcal{L}^{MULT}$ is maximized and the adversarial loss $\mathcal{L}^{adv}$ of the adversarial module $h(\cdot)$ is minimized as in Equation~\ref{eq:minmax}.

\begin{equation}
    \label{eq:minmax}
    \underset{f,g}{\arg max}\ \underset{h}{\arg min} \ \mathcal{L}^{MULT}(x)- \mathcal{L}^{adv}(x,y)
\end{equation}

In order to make the adversarial training possible through a simple backpropagation process, a gradient reversal layer $grl(\cdot)$ \cite{pmlr-v37-ganin15} is added to the adversarial module. The purpose of $grl(\cdot)$ is to scale and change the direction of the calculated gradients using a gradient reversal scaling factor $\lambda$ in the backward pass during training, and work as the identity function during the forward pass. This addition also allows the reformulation of the learning objective, as shown in Equation~\ref{eq:minmaxref}.
\begin{equation}
    \label{eq:minmaxref}
    \underset{f,g,h}{\arg min}\ \mathcal{L}^{MULT}(x)+ \mathcal{L}^{adv}(x,y)
\end{equation}
The attack phase of the \modeladvmultvae starts after the optimization of the previously introduced learning objective~(Equation~\ref{eq:minmaxref}). In this phase, the resultant trained model, with frozen parameters, is used to train a standalone attacker network $h^{atk}(\cdot)$ aiming to decipher the remaining sensitive information encoded in the latent vector $z$ generated by the encoder $f(\cdot)$.  
\begin{figure}
\begin{center}
    \includegraphics[width=0.65\textwidth]{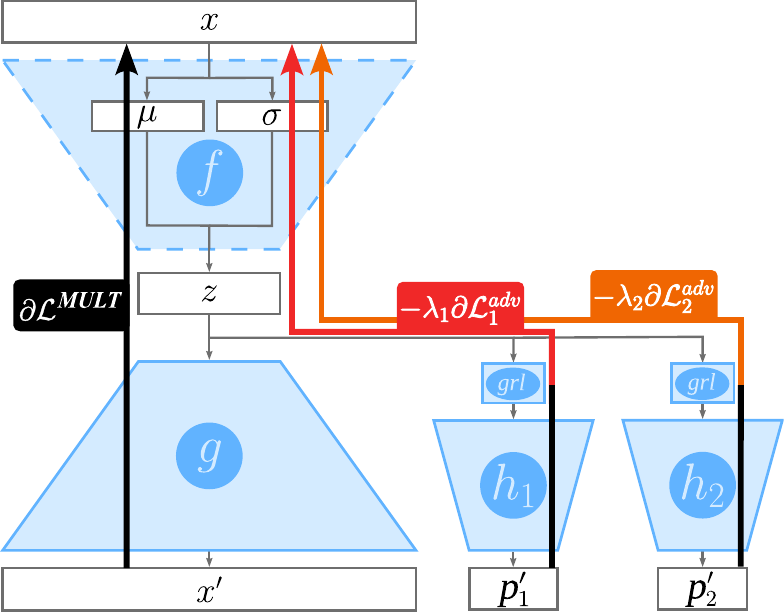}
    \caption{Outline of our \modelmadvmultvae approach. The thin gray arrows flowing from the top to the bottom indicate the forward pass, the bold arrows the backward pass. Here, the red and orange arrows highlight the reversed gradients of the attributes, $p_1$ and $p_2$,  respectively, that the model should unlearn.}
    \label{fig:advxmultvae}
\end{center}
\end{figure}

The architecture of our proposed \modelmadvmultvae approach for simultaneous removal of attributes is illustrated in Figure~\ref{fig:advxmultvae}.
Formally, we first define the set $\mathcal{P}:\{p_0, p_1, \ldots, p_k\}$ representing the users' $k$ protected attributes, where each $p_i$, $i \in \{0, \ldots, k\}$, can be of a continuous or categorical type. To unlearn each of these attributes, instead of the single adversarial module in \modeladvmultvae, we define a set of corresponding adversarial modules $\mathcal{H}:\{h_0,\ldots,h_k\}$. Each of the modules receives $z$ as input and aims at predicting the protected attribute $p_k$. Therefore, we define the loss of each module as $\mathcal{L}^{adv}_k(h_k(z),p_k)$ and use cross-entropy (CE) for categorical attributes and mean-squared-error (MSE) for continuous attributes. Given these definitions, we introduce the multi-attribute adversarial loss $\mathcal{L}^{advX}$ (Equation~\ref{eq:adxloss}) as the sum of each of the losses assigned to each adversarial module $h_k$, and integrate it into the final loss, shown in Equation~\ref{eq:adxtotal}. 

\begin{equation}
 \label{eq:adxloss}
    \mathcal{L}^{advX} = \sum_{k}^{|\mathcal{P}|} \mathcal{L}^{adv}_k(h_k(z),p_k)  
\end{equation}

\begin{equation}
    \label{eq:adxtotal}
    \underset{f,g,\{h_0,\ldots,h_k\}}{\arg min}\ \mathcal{L}^{MULT}(x)+ \mathcal{L}^{advX}(x,\mathcal{P})
\end{equation}

With this formulation, we are also able to reproduce the previously introduced models by setting adequate values of $\lambda$s for each module. 

\section{Experimental Setup}\label{sec:experimental}
\paragraph{Datasets.}
We derived \dataLFM from \dataorigLFM by only considering users for which we have gender and age information. We then 
subsample 100k tracks uniformly at random and reinforce 10-core filtering. For \dataMlm we adopt 5-core filtering. Table~\ref{tab:datasets} describes the resultant datasets. 
Furthermore, the users' age information was normalized by scaling the original values in the range $\left[0:60\right]$ for \dataMlm and $\left[0:120\right]$ for \dataLFM to cover their all their possible values. The resultant dataset is randomly split in 5-folds across users. For each fold, we randomly select $20\%$ of users as test slice, then, the $20\%$ of the remaining users are selected for validation, and the rest for training.
We mitigate the imbalanced gender distribution by adding proportional weights to the CE loss of the corresponding adversarial module. 

\begin{table}[t]
    \centering
    \caption{Statistical description of datasets}
\begin{tabular}{lrrrrr}
\toprule
Dataset&\multicolumn{2}{r}{Users}     &Items & Interactions & Density \\\midrule
{\multirow{4}{*}{\dataMlm}}
&Total & 6,040 &3,416 &999,611 & $\quad$ 0.0484\\
&Gender (Male/Female) &4,331/1,709 &&&\\   
&Age (Mean/Std/Median)&30.6/12.9/25.0&&&\\\midrule

{\multirow{4}{*}{\dataLFM}}
&Total & 8,663 & 47,383 & 1,476,788  & 0.0036\\
&Gender (Male/Female) &7,012/1,651&&&   \\
&Age (Mean/Std/Median)& 25.4/8.4/23.0 &&&\\
\bottomrule

%
\end{tabular}

\label{tab:datasets}
\end{table}

\paragraph{Evaluation.}
We adopt the \textit{normalized discounted cumulative gain}~(\ndcg)~\cite{kalervo2002ndcg} and \Recall metrics to asses the accuracy of recommendation performance for the top 10 recommended items.  
In addition, we measure the attacker networks' ability to predict protected attributes and we report the corresponding \textit{balanced accuracy} ($BAcc$) \cite{DBLP:conf/icpr/BrodersenOSB10} for categorical attributes, which values $\sim\frac{1}{\#categories}$ indicate total debiasing of the predicted attribute, therefore, the inability of the attacker network to successfully predict the given attribute~(\eg $BAcc=0.5$ for binary classification). Moreover, we report the \textit{mean absolute error} ($MAE$) for continuous attributes, which higher values indicate a stronger debiasing effect achieved. 

 We report the average and standard deviation for the test slices across 5-folds. In addition, we concatenate the users' individual scores in all folds and then perform Wilcoxon signed-rank tests~\cite{Rey2011} to asses the statistical significance differences between recommendation models, McNemar's tests~\cite{mcnemar1947note} to measure the agreement of results of gender attacker networks, and t-tests between age attacker networks' results. All statistical tests are performed with $95\%$ of confidence between the results of different (\lambgender, \lambage) combinations. 

\paragraph{Models and training procedure.}
We train our models inspired by the protocol presented in \cite{10.1145/3477495.3531820}. Consequently, for each dataset, we target the single and simultaneous removal of \textit{age}~(continuous) and \textit{gender}~(categorical).  
We fix the following parameters to establish fair conditions of comparison between models: We train all models for 200 epochs for the adversarial removal phase and 50 epochs for the attack phase for both datasets using an Adam optimizer. The adversarial and attacker networks have the same architecture represented by a multi-layer perceptron with one intermediate layer, and according to the type of attribute, one single value for a continuous attribute or a one-hot encoded vector 
for a categorical attribute is predicted. For the attacking phase, we avoid the sampling step used during the training of the \modelmultvae and use the direct output of the encoder $f(\cdot)$ for each user. For each model, we perform an exhaustive grid search to identify optimized combinations of gradient reversal scaling factors \lambgender (gender) and \lambage (age), to observe the effect on performance and debiasing for different degrees of removal aggressiveness. We investigate $\lambda$ values\footnote{We include $\lambda$=1 for maximal removal as in the original formulation~\cite{pmlr-v37-ganin15} } in $\left[0,\ 1,\ 200,\ 400,\ 600,\ 800\right]$ due to the difference in magnitude between   $\mathcal{L}^{MULT}$ and  $\mathcal{L}^{advX}$ that we devised in early experiments. We report results for the best debiasing results of each attribute and present an analysis of all 36 different (\lambgender, \lambage) combinations.  
The training of the attacker networks is performed regardless of the values of \lambgender~and \lambage. The source code of and complete configuration of our experiments can be found in our repository.\footnote{\url{https://github.com/hcai-mms/advx-multvae}} 
\begin{table}[t]
\caption{Experimental results expressed in percentages on the two datasets \dataMlm and \dataLFM. The scores in \textbf{bold} indicate the best scores across all models. The superscript $\spadesuit$ indicates statistical significance difference with $p\le0.05$ between \modeladvmultvae and \modelmadvmultvae for the given attribute in the suffix of the models' names. The superscript $\star$ indicates statistical significance difference with $p\le0.05$ across all models.}
    \centering
    
    \setlength{\tabcolsep}{3pt} 
    \renewcommand{\arraystretch}{1} 
    \begin{tabular}{c l  c c c c c}
    \toprule
   & & \multicolumn{2}{c}{Performance}&\multicolumn{2}{c}{Debiasing} \\ 
   \cmidrule(lr){3-4}\cmidrule{5-6}
    Dataset&Model-Attribute& $NDCG\uparrow$ & $Recall\uparrow$ & \baccgender$\downarrow$ & \maeage$\uparrow$ \\ \midrule
   
 {\multirow{5}{*}{\dataMlm}} 
&\modelmultvae & $\mathbf{62.72_{0.98}}^\star$ &     $\mathbf{08.16_{0.09}}$ & $69.81_{1.82}$& $14.45_{0.38}$ \\\cdashline{2-7}
&\modeladvmultvaegen&       $61.15_{1.22}$ &     $07.84_{0.13}$ & $59.05_{5.57}$ & $14.52_{0.47}$  \\
&\modelmadvmultvaegen&      $61.21_{1.11}$ &     $07.88_{0.10}$ & $\mathbf{57.14_{5.72}}^{\spadesuit\star}$ & $16.91_{0.39}$  \\\cdashline{2-7}
&\modeladvmultvaeage&       $62.51_{1.05}^\spadesuit$ &     $08.12_{0.04}^\spadesuit$ & $69.33_{1.58}$ & $\mathbf{17.10_{0.42}}$ \\

&\modelmadvmultvaeage&      $61.05_{1.16}$ &     $07.86_{0.10}$ & $59.02_{3.82}$ & $17.09_{0.43}$  \\
\midrule
{\multirow{5}{*}{\dataLFM}} 
&  \modelmultvae            &    $\mathbf{47.46_{0.24}}$ &  $06.54_{0.22}$ &  $67.00_{1.43}$ & $04.28_{0.16}$ \\\cdashline{2-7}
&  \modeladvmultvaegen      &    $46.52_{0.23}^\spadesuit$ &  $06.41_{0.12}^\spadesuit$ &  $53.03_{2.51}$ & $04.32_{0.18}$ \\
&  \modelmadvmultvaegen     &    $45.88_{0.29}$ &  $06.27_{0.16}$ &  $\mathbf{52.81_{1.79}}^{\spadesuit\star}$ & $04.33_{0.16}$ \\
\cdashline{2-7}
&  \modeladvmultvaeage      &    $47.42_{0.34}^\spadesuit$ &  $\mathbf{06.59_{0.12}}^\spadesuit$ &  $68.02_{1.38}$ & $04.37_{0.13}$ \\

&  \modelmadvmultvaeage     &    $46.77_{0.28}$ &  $06.45_{0.17}$ &  $53.37_{1.63}$ & $\mathbf{04.40_{0.16}}^{\spadesuit\star}$ \\
     \bottomrule
    \end{tabular}
    \label{tab:results}
\end{table}

\section{Results \& Analyses}\label{sec:results}
In this Section, we first analyze the overall performance of the investigated models and then we present a analysis on the multiple debiasing objective and the impact of different  (\lambgender, \lambage) combinations on debiasing and recommendation performance for the \dataMlm dataset.  
\subsection{Overall Performance}
\Tabref{tab:results} shows the performance of various models and attacker networks. We split the results in three groups according to the explored attribute, identified by the suffixes \texttt{\textsc{G}}~(gender) and \texttt{\textsc{A}}~(age). Each row represents the best debiasing results in percentages for the attribute in the suffix of each model's name. Each score's subscript indicate the standard deviation across 5-folds. 

In the Table~\ref{tab:results}, the removal of gender (\modeladvmultvaegen and \modelmadvmultvaegen) for both datasets has similar improvements in terms of \baccgender. However, when targeting the removal of age (\modeladvmultvaeage and \modelmadvmultvaeage), we obtain a substantial improvement in terms of \maeage for the \dataMlm dataset and are very similar \maeage values for the \dataLFM dataset, which might reflect the differences of user's age values distributions between the two studied datasets, and also a lower correlation between users preferences and their declared age in the LastFM platform which we take a limitation for this work.

For \modeladvmultvae, we observe that targeting gender removal causes a $\sim$3\% drop in \ndcg and \Recall for both datasets, as expected~\cite{10.1145/3477495.3531820}, compared to \modelmultvae showing a substantial drop in \baccgender of $\sim$15-20\%. The age removal, 
 leads to a $\sim$17\% increase of \maeage for \dataMlm and remains the same for \dataLFM. 
Also, \modeladvmultvaeage and \modeladvmultvaegen's results show that the single removal of gender has a slight indirect effect on the removal of age and vice-versa. We speculate that the difference in the removal effect might be due to the difference in magnitude of the gradients that the MSE an CE adversarial networks objective yield.

Our approach \modelmadvmultvae obtains similar \ndcg values and shows the best mean values of \baccgender with a slight difference in \maeage for both datasets in comparison to \modeladvmultvae and presents an overlap if we examine their standard deviation values (\eg \baccgender values for \modeladvmultvaegen and \modelmadvmultvaegen). Nonetheless, except for \baccgender of \modelmadvmultvaegen, all the best reported debiasing metrics are statistically significant. Furthermore, \modelmadvmultvae's results indicate a trade-off between \baccgender and  \maeage, when we target the best removal of each attribute on both datasets. However, this trade-off does not have a great impact across all examined metrics.

\subsection{Multiple Debiasing Objective Behavior}
\begin{figure}[t]
 \centering
  \includegraphics[width=0.75
\linewidth]{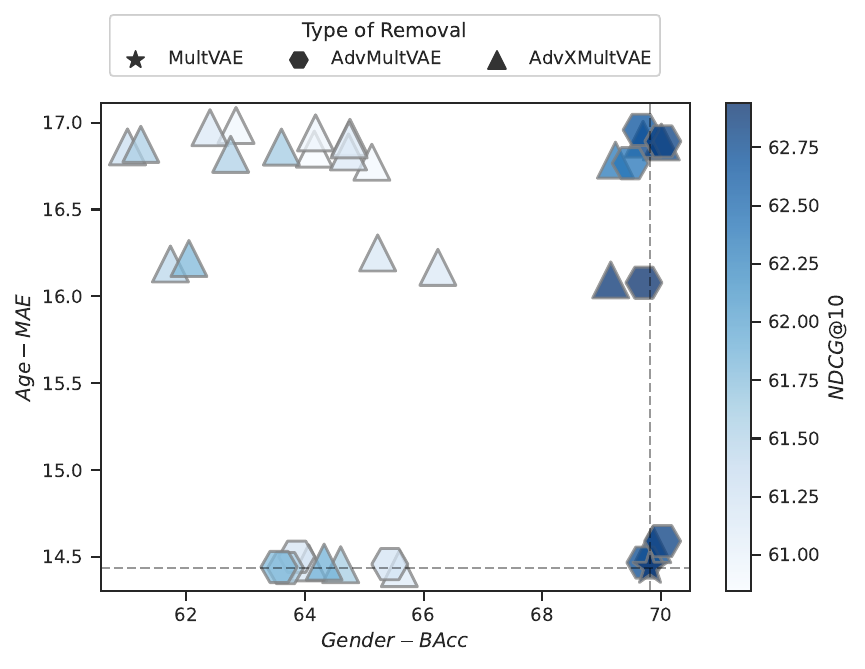}%
\caption{Performance for debiasing (\baccgender and \maeage) and recommendation accuracy ($NDCG@10$) for the \dataMlm dataset. Points on the upper-left corner refer to the best privacy-preserving models where the \modelmadvmultvae removal variants yield the strongest debiasing power with a marginal loss in terms of \ndcg. The intersection of the dotted lines represent the model without debiasing (\modelmultvae).}

\label{fig:ml-dist-multiobj}
\end{figure}

To delve into the behavior all the explored (\lambgender, \lambage) combinations and the multiple debiasing objective behavior, Figure~\ref{fig:ml-dist-multiobj} shows all values of debiasing metrics for the \dataMlm dataset with their corresponding \ndcg scores for all (\lambgender,~\lambage) pairs.
Each point of the distribution represents one (\lambgender,~\lambage) combination and the different models are represented by different shapes. We also include dotted lines which represent the debiasing scores obtained for \modelmultvae as reference for the analysis.

The distribution of points in Figure~\ref{fig:ml-dist-multiobj} illustrates the trade-off between debiasing power and recommendation performance, where the models in the upper-left region have stronger debiasing capabilities and higher \ndcg values in the lower-right corner. Moreover, the variants to the right of the vertical dotted line indicate higher \baccgender values than \modelmultvae but obtain similar \ndcg values (we explore this effect in more detail in Subsection~\ref{subsec:inter_lamb}). Furthermore, the \modeladvmultvae variants are in the vicinity of the dotted lines which indicates a mild influence on the debiasing of one attribute onto the other, this also can be observed in Table \ref{tab:results}.

Our approach \modelmadvmultvae can achieve stronger debiasing capabilities than their \modeladvmultvae counterparts with a marginal drop in recommendation performance for both attributes, and especially for gender. This is  indicated by the presence of  \modelmadvmultvae points on the upper-left corner and towards the direction of our multiple debiasing objective. Also, the corresponding performance values in this region are not drastically lower than those of the \modeladvmultvae and \modelmultvae variants. 
Additionally, some \modelmadvmultvae removal variants on the upper-left corner show slightly better \ndcg than those in the upper-middle region, which indicates that more aggressive simultaneous removal can yield higher accuracy in some cases. We observe that it is possible to realize age removal to a substantial extent with only a marginal drop in \ndcg, which is indicated by the removal variants in the upper-right corner. Furthermore some \modelmadvmultvae variants lay close to the dotted lines and correspond to the configurations when 
\lambgender~or \lambage~is set to $1$ which results in similar behavior to the \modeladvmultvae variants.

\begin{figure}[ht]
\begin{subfigure}{.33\textwidth}
  \centering
\includegraphics[width=\linewidth]{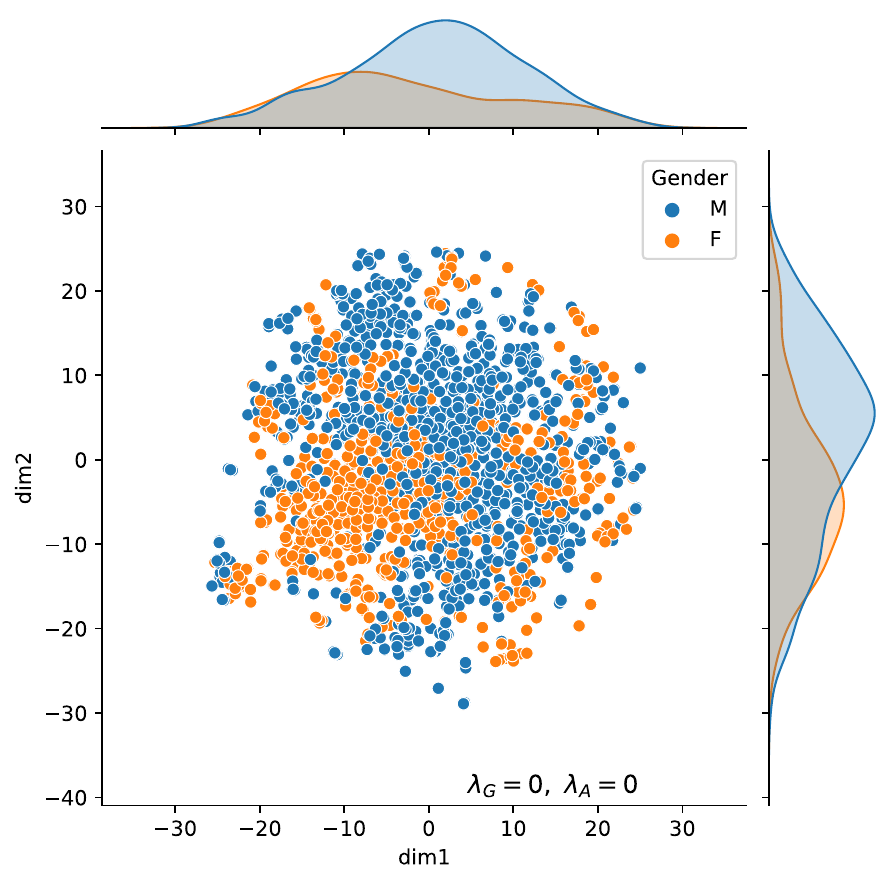}
  \caption{\modelmultvae}
  \label{fig:tsne00}
\end{subfigure}%
\begin{subfigure}{.33\textwidth}
  \centering
\includegraphics[width=\linewidth]{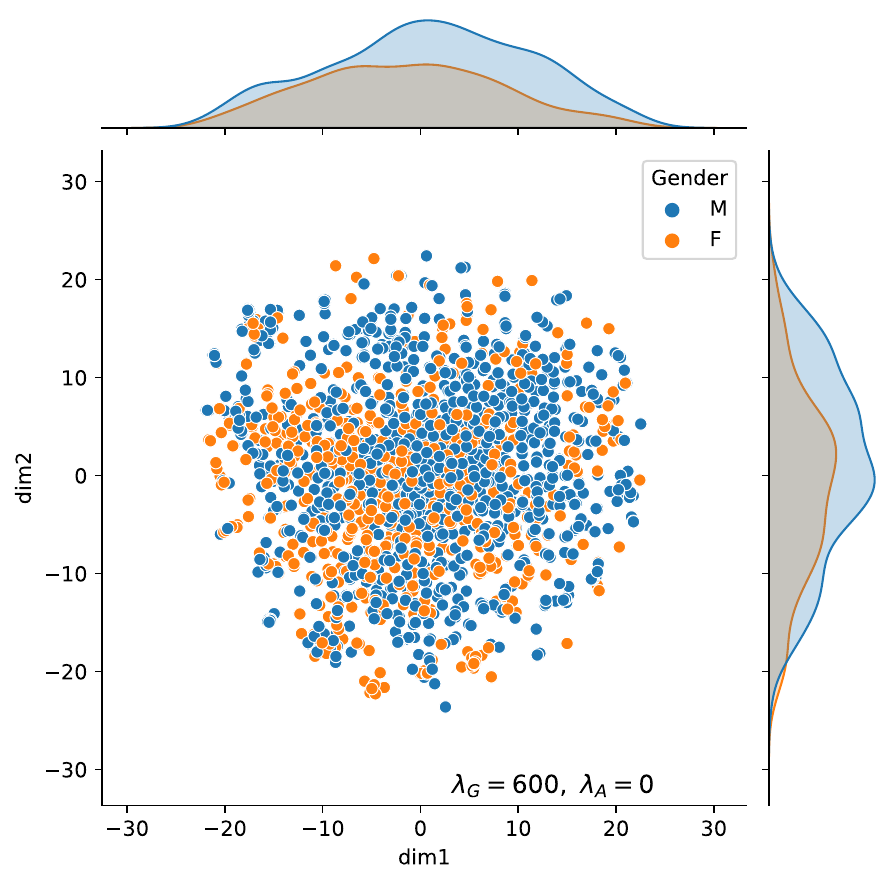}
  \caption{\modeladvmultvae}
  \label{fig:tsne6000}
\end{subfigure}
\begin{subfigure}{.33\textwidth}
  \centering
\includegraphics[width=\linewidth]{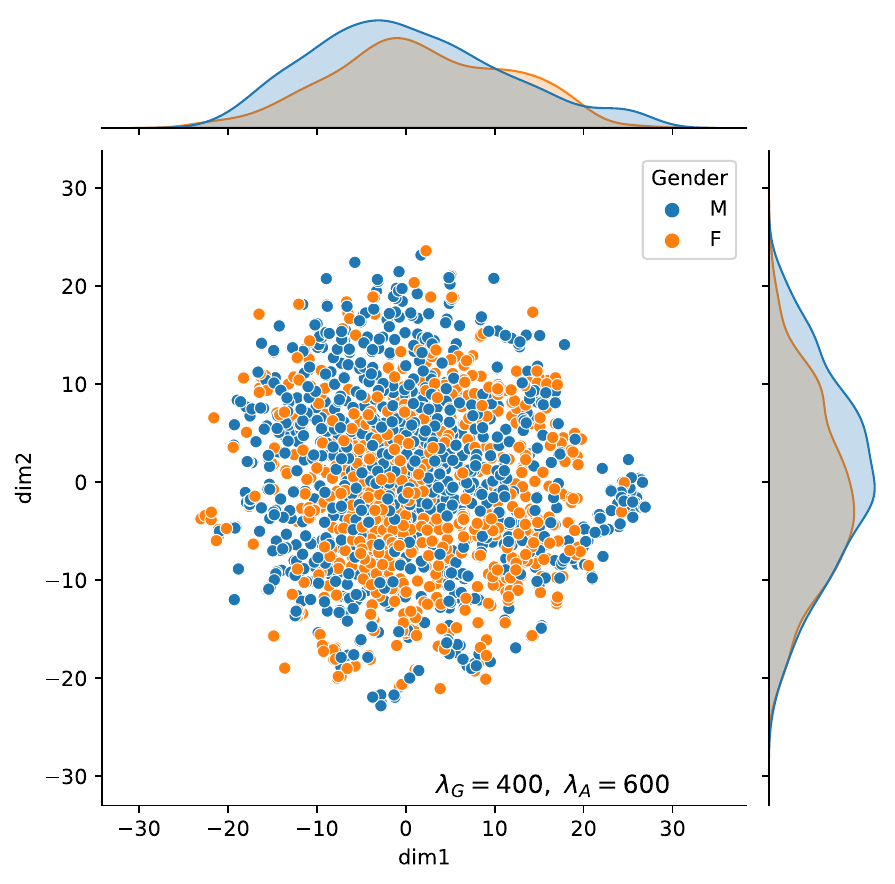}
  \caption{\modelmadvmultvae}
  \label{fig:tsne400600}
\end{subfigure}

\caption{t-SNE plots of the users' latent embeddings and attacker networks' predictions of gender for the \dataMlm dataset, each plot includes the density of the distribution for each gender across the corresponding latent dimension.}
\label{fig:tsne-ml-gender}
\end{figure}

In order to show the ability of the attacker networks to map real the distribution of the users' gender, 
we show t-SNE~\cite{van2008visualizing} plots of the first and second dimensions of the user's latent embeddings of our three variants for the \dataMlm dataset in the Fig.~\ref{fig:tsne-ml-gender}. Alongside each plot's axes, we show the corresponding density of the distributions of predicted gender. Figure~\ref{fig:tsne00} shows a distribution of the two predicted classes for the \modeladvmultvae model where the densities of the two classes are visibly discordant . In contrast, we observe that the distributions are less discordant for the \modeladvmultvae~(Fig.~\ref{fig:tsne6000}) and \modelmadvmultvae~(Fig.~\ref{fig:tsne400600}) variants which indicates that is harder to infer gender information after applying adversarial removal. Additionally, our approach \modelmadvmultvae~(Fig.~\ref{fig:tsne400600}), shows the most overlap between distributions of gender, therefore, yields the most privacy-preserving user's latent embeddings. 

\subsection{Interaction of Multiple Gradient Reversal Scaling Factors}\label{subsec:inter_lamb}
\begin{figure}[h]
    \centering
 \centering
  \includegraphics[width=\linewidth]{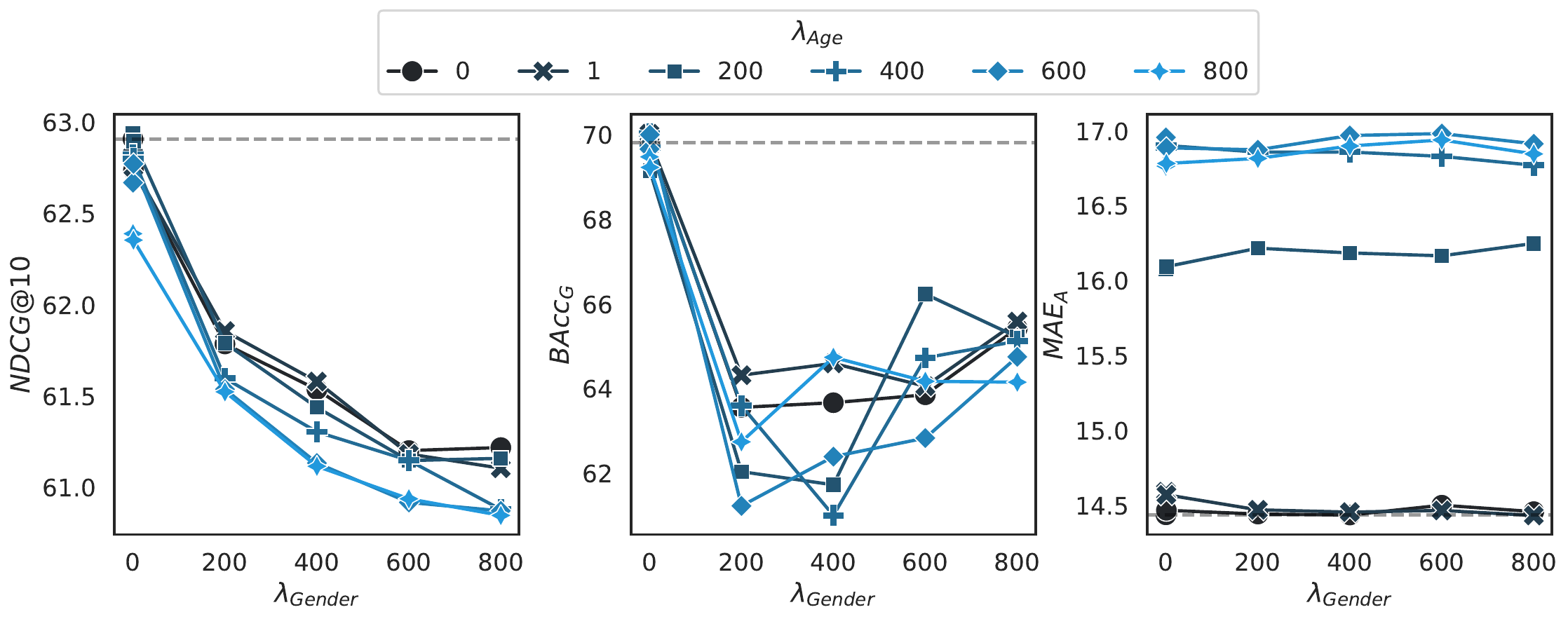}
\caption{Interaction of different gradient scaling factors (\lambgender, \lambage)  and reported metrics for the \dataMlm dataset. Each subplot indicates the obtained mean values 
 of \ndcg, \baccgender, and \maeage respectively from left to right.}
 \label{fig:ml_lamb-inter}
\end{figure}

In order to observe the interaction of the explored values of lambdas, Figure \ref{fig:ml_lamb-inter} shows how these values affect debiasing and recommendation performance for the \dataMlm dataset. We set the \lambgender~values in the $x$-axis and the average across 5-folds of \ndcg~(\textit{left}), \baccgender~(\textit{middle}), and \maeage~(\textit{right}) as the $y$-axis for each subplot respectively. Each subplot shows different lines corresponding to the studied \lambage~values. The dotted lines represent the scores for \modelmultvae as a reference for the analysis.

In the \textit{left} subplot, we observe a consistent decrease of the \ndcg values when we increase the both values of lambda. More specifically, when only gender is removed (\lambage$=0$) we obtain higher \ndcg values across the different \lambgender~ values, which indicates the trade-off of performance and debiasing of more than one attribute. Also, when \lambgender$\sim1$, we can even achieve slightly higher \ndcg than \modelmultvae's  for \lambage$=200$.  

In the \textit{middle} subplot, a pronounced \baccgender drop when \lambgender$\ge=200$ for all the \lambage~values, and especially for \lambage$=600$ and \lambage$=200$ where we can see the positive effect of simultaneously removing both attributes. Moreover, we observe a negative impact to the debiasing of gender presenting an increasing tendency of \baccgender values when \lambgender$\ge 600$. Also, the best removal of gender is observed when \lambgender$=400$ and \lambage$=400$. Moreover, when \lambgender$\sim1$, we can even achieve slightly worse \baccgender than \modelmultvae's results, which was also observed in Fig.~\ref{fig:ml-dist-multiobj}.     

In the \textit{right} subplot, the values of \maeage present slight fluctuations across the different values of \lambgender, and a more pronounced debiasing effect when \lambage$\ge200$. Moreover,  when \lambage$\ge400$, we obtain similar \maeage values, which might indicate a removal saturation point for \lambage. Furthermore, the best \maeage value is achieved when \lambgender$=600$ and \lambage$=600$. 

Overall, from Figure~\ref{fig:ml_lamb-inter}, we can establish that using high values of \lambage~and \lambgender~leads to a consistent drop of \ndcg. Additionally, using high values of \lambgender~has a detrimental effect on the debiasing of gender which we speculate to be an effect of steering the encoder updates to the objective of the adversarial modules during training. Also, the multiple \lambgender~explored values do not influence the removal of age which might indicate low dependency between these two private attributes.    

\section{Conclusion}
In this work, we presented \modelmadvmultvae to address the simultaneous removal of users' multiple protected attributes of continuous and categorical types, using adversarial training in a variational autoencoder architecture. Our results show that the simultaneous removal of gender and age can yield stronger debiasing capabilities to its single-attribute removal counterparts with a slight drop in recommendation performance. Further work could explore the debiasing of more than two attributes, for which our architecture can easily be extended, as well as experimenting with other attributes, e.g.~personality traits. Also, the simultaneous removal of attributes could be applied to other representations, such as item embeddings (to unlearn music artist's gender information, for instance), as well as recommendation algorithms beyond VAEs. 

\begin{credits}
\subsubsection{\ackname} 
This research was funded in whole or in part by the Austrian Science Fund (FWF): P36413, P33526, and DFH-23, and by the State of Upper Austria and the Federal Ministry of Education, Science, and Research, through grants LIT-2021-YOU-215 and LIT-2020-9-SEE-113.
\subsubsection{\discintname}
The authors have no competing interests to declare that are relevant to the content of this article.
\end{credits}
%
%
%

\begin{thebibliography}{10}
\providecommand{\url}[1]{\texttt{#1}}
\providecommand{\urlprefix}{URL }
\providecommand{\doi}[1]{https://doi.org/#1}

\bibitem{beigi2020survey}
Beigi, G., Liu, H.: A survey on privacy in social media: Identification, mitigation, and applications. ACM/IMS Trans. Data Sci.  \textbf{1}(1) (mar 2020). \doi{10.1145/3343038}, \url{https://doi.org/10.1145/3343038}

\bibitem{DBLP:conf/icpr/BrodersenOSB10}
Brodersen, K.H., Ong, C.S., Stephan, K.E., Buhmann, J.M.: The balanced accuracy and its posterior distribution. In: 20th International Conference on Pattern Recognition, {ICPR} 2010, Istanbul, Turkey, 23-26 August 2010. pp. 3121--3124. {IEEE} Computer Society (2010). \doi{10.1109/ICPR.2010.764}, \url{https://doi.org/10.1109/ICPR.2010.764}

\bibitem{DBLP:journals/csur/DeldjooNM21}
Deldjoo, Y., Noia, T.D., Merra, F.A.: A survey on adversarial recommender systems: From attack/defense strategies to generative adversarial networks. {ACM} Comput. Surv.  \textbf{54}(2),  35:1--35:38 (2022). \doi{10.1145/3439729}, \url{https://doi.org/10.1145/3439729}

\bibitem{10.1145/3477495.3531820}
Ganh\"{o}r, C., Penz, D., Rekabsaz, N., Lesota, O., Schedl, M.: Unlearning protected user attributes in recommendations with adversarial training. In: Proceedings of the 45th International ACM SIGIR Conference on Research and Development in Information Retrieval. p. 2142–2147. SIGIR '22, Association for Computing Machinery, New York, NY, USA (2022). \doi{10.1145/3477495.3531820}, \url{https://doi.org/10.1145/3477495.3531820}

\bibitem{pmlr-v37-ganin15}
Ganin, Y., Lempitsky, V.: Unsupervised domain adaptation by backpropagation. In: Bach, F., Blei, D. (eds.) Proceedings of the 32nd International Conference on Machine Learning. Proceedings of Machine Learning Research, vol.~37, pp. 1180--1189. PMLR, Lille, France (07--09 Jul 2015), \url{https://proceedings.mlr.press/v37/ganin15.html}

\bibitem{hauzenberger-etal-2023-modular}
Hauzenberger, L., Masoudian, S., Kumar, D., Schedl, M., Rekabsaz, N.: Modular and on-demand bias mitigation with attribute-removal subnetworks. In: Findings of the Association for Computational Linguistics: ACL 2023. pp. 6192--6214. Association for Computational Linguistics, Toronto, Canada (Jul 2023). \doi{10.18653/v1/2023.findings-acl.386}, \url{https://aclanthology.org/2023.findings-acl.386}

\bibitem{kalervo2002ndcg}
J\""{a}rvelin, K., Kek\"{a}l\"{a}inen, J.: Cumulated gain-based evaluation of ir techniques. ACM Trans. Inf. Syst.  \textbf{20}(4),  422–446 (oct 2002). \doi{10.1145/582415.582418}, \url{https://doi.org/10.1145/582415.582418}

\bibitem{kumar-etal-2023-parameter}
Kumar, D., Lesota, O., Zerveas, G., Cohen, D., Eickhoff, C., Schedl, M., Rekabsaz, N.: Parameter-efficient modularised bias mitigation via {A}dapter{F}usion. In: Proceedings of the 17th Conference of the European Chapter of the Association for Computational Linguistics. pp. 2738--2751. Association for Computational Linguistics, Dubrovnik, Croatia (May 2023). \doi{10.18653/v1/2023.eacl-main.201}, \url{https://aclanthology.org/2023.eacl-main.201}

\bibitem{Li2021counterFactMult}
Li, Y., Chen, H., Xu, S., Ge, Y., Zhang, Y.: Towards personalized fairness based on causal notion. In: Proceedings of the 44th International ACM SIGIR Conference on Research and Development in Information Retrieval. p. 1054–1063. SIGIR '21, Association for Computing Machinery, New York, NY, USA (2021). \doi{10.1145/3404835.3462966}, \url{https://doi.org/10.1145/3404835.3462966}

\bibitem{LiangMultVAE2018}
Liang, D., Krishnan, R.G., Hoffman, M.D., Jebara, T.: Variational autoencoders for collaborative filtering. In: Proceedings of the 2018 World Wide Web Conference. p. 689–698. WWW '18, International World Wide Web Conferences Steering Committee, Republic and Canton of Geneva, CHE (2018). \doi{10.1145/3178876.3186150}, \url{https://doi.org/10.1145/3178876.3186150}

\bibitem{adv-LiuWLXZ22}
Liu, H., Wang, Y., Lin, H., Xu, B., Zhao, N.: Mitigating sensitive data exposure with adversarial learning for fairness recommendation systems. Neural Comput. Appl.  \textbf{34}(20),  18097--18111 (2022). \doi{10.1007/s00521-022-07373-4}, \url{https://doi.org/10.1007/s00521-022-07373-4}

\bibitem{van2008visualizing}
Van~der Maaten, L., Hinton, G.: Visualizing data using t-sne. Journal of machine learning research  \textbf{9}(11) (2008)

\bibitem{mcnemar1947note}
McNemar, Q.: Note on the sampling error of the difference between correlated proportions or percentages. Psychometrika  \textbf{12}(2),  153--157 (1947)

\bibitem{MELCHIORRE2021102666}
Melchiorre, A.B., Rekabsaz, N., Parada-Cabaleiro, E., Brandl, S., Lesota, O., Schedl, M.: Investigating gender fairness of recommendation algorithms in the music domain. Information Processing \& Management  \textbf{58}(5),  102666 (2021). \doi{10.1016/j.ipm.2021.102666}, \url{https://www.sciencedirect.com/science/article/pii/S0306457321001540}

\bibitem{Rey2011}
Rey, D., Neuh{\"a}user, M.: Wilcoxon-Signed-Rank Test, pp. 1658--1659. Springer Berlin Heidelberg, Berlin, Heidelberg (2011). \doi{10.1007/978-3-642-04898-2\_616}, \url{https://doi.org/10.1007/978-3-642-04898-2_616}

\bibitem{10.1145/3511047.3536400}
Schedl, M., Rekabsaz, N., Lex, E., Grosz, T., Greif, E.: Multiperspective and multidisciplinary treatment of fairness in recommender systems research. In: Adjunct Proceedings of the 30th ACM Conference on User Modeling, Adaptation and Personalization. p. 90–94. UMAP '22 Adjunct, Association for Computing Machinery, New York, NY, USA (2022). \doi{10.1145/3511047.3536400}, \url{https://doi.org/10.1145/3511047.3536400}

\bibitem{Wang2023TrustRec}
Wang, S., Zhang, X., Wang, Y., Ricci, F.: Trustworthy recommender systems. ACM Trans. Intell. Syst. Technol.  (oct 2023). \doi{10.1145/3627826}, \url{https://doi.org/10.1145/3627826}, just Accepted

\bibitem{Wang23SurveyFairness}
Wang, Y., Ma, W., Zhang, M., Liu, Y., Ma, S.: A survey on the fairness of recommender systems. ACM Trans. Inf. Syst.  \textbf{41}(3) (feb 2023). \doi{10.1145/3547333}, \url{https://doi.org/10.1145/3547333}

\bibitem{Wang2022InferenceAttack}
Wang, Z., Huang, N., Sun, F., Ren, P., Chen, Z., Luo, H., de~Rijke, M., Ren, Z.: Debiasing learning for membership inference attacks against recommender systems. In: Proceedings of the 28th ACM SIGKDD Conference on Knowledge Discovery and Data Mining. p. 1959–1968. KDD '22, Association for Computing Machinery, New York, NY, USA (2022). \doi{10.1145/3534678.3539392}, \url{https://doi.org/10.1145/3534678.3539392}

\end{thebibliography}
%

\end{document}